\documentclass[letterpaper]{article} % DO NOT CHANGE THIS
\usepackage{aaai25}  % DO NOT CHANGE THIS
\usepackage{times}  % DO NOT CHANGE THIS
\usepackage{helvet}  % DO NOT CHANGE THIS
\usepackage{courier}  % DO NOT CHANGE THIS
\usepackage[hyphens]{url}  % DO NOT CHANGE THIS
\usepackage{graphicx} % DO NOT CHANGE THIS
\usepackage{natbib}  % DO NOT CHANGE THIS AND DO NOT ADD ANY OPTIONS TO IT
\usepackage{caption} % DO NOT CHANGE THIS AND DO NOT ADD ANY OPTIONS TO IT
\usepackage{algorithm}
\usepackage{algorithmic}

\usepackage{newfloat}
\usepackage{listings}
\DeclareCaptionStyle{ruled}{labelfont=normalfont,labelsep=colon,strut=off} % DO NOT CHANGE THIS
\floatstyle{ruled}
\newfloat{listing}{tb}{lst}{}
\floatname{listing}{Listing}
\usepackage{subfigure}
\usepackage{subcaption}
\usepackage{amsmath,amssymb,amsthm}
\usepackage{multirow}
\usepackage{arydshln}
\usepackage{booktabs}
\usepackage{color}
\usepackage{amsfonts}
\usepackage{nicefrac}
\usepackage{microtype}
\usepackage{xcolor}
\usepackage{caption}
\usepackage{lipsum}
\usepackage{placeins}

\title{VersaGen: Unleashing Versatile Visual Control for Text-to-Image Synthesis}
\author{
    Zhipeng Chen\textsuperscript{\rm 1}, Lan Yang\textsuperscript{\rm 1, \rm 2}, Yonggang Qi\textsuperscript{\rm 1, \rm 2},\\
    Honggang Zhang\textsuperscript{\rm 1}, Kaiyue Pang\textsuperscript{\rm 2}, Ke Li\textsuperscript{\rm 1, \rm 2}\thanks{Corresponding author}, Yi-Zhe Song\textsuperscript{\rm 2}
}
\affiliations{
    \textsuperscript{\rm 1}School of Artificial Intelligence, Beijing University of Posts and Telecommunications, China\\
    \textsuperscript{\rm 2}SketchX, CVSSP, University of Surrey, United Kingdom\\
    zhipengchen1998@bupt.edu.cn, ylan@bupt.edu.cn, qiyg@bupt.edu.cn, \\
    zhhg@bupt.edu.cn, thatkpang@gmail.com, like1990@bupt.edu.cn, y.song@surrey.ac.uk
}

\usepackage{bibentry}
\begin{document}

\maketitle

\begin{abstract}
Despite the rapid advancements in text-to-image (T2I) synthesis, enabling precise visual control remains a significant challenge. Existing works attempted to incorporate multi-facet controls (text and sketch), aiming to enhance the creative control over generated images. However, our pilot study reveals that the expressive power of humans far surpasses the capabilities of current methods. Users desire a more versatile approach that can accommodate their diverse creative intents, ranging from controlling individual subjects to manipulating the entire scene composition. We present VersaGen, a generative AI agent that enables versatile visual control in T2I synthesis. VersaGen admits four types of visual controls: i) single visual subject; ii) multiple visual subjects; iii) scene background; iv) any combination of the three above or merely no control at all. We train an adaptor upon a frozen T2I model to accommodate the visual information into the text-dominated diffusion process. We introduce three optimization strategies during the inference phase of VersaGen to improve generation results and enhance user experience. Comprehensive experiments on COCO and Sketchy validate the effectiveness and flexibility of VersaGen, as evidenced by both qualitative and quantitative results. 
\begin{links}
  \link{Code}{https://github.com/FelixChan9527/VersaGen_official}
\end{links}
\end{abstract}

\begin{figure}[htb]
    \centering
    \includegraphics[width=\columnwidth]{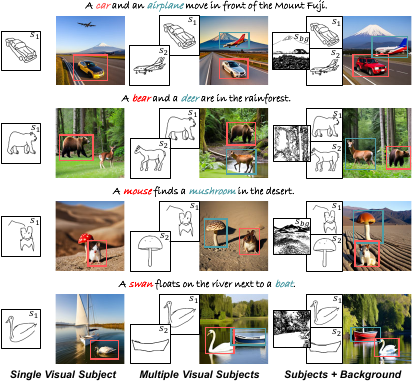}
    \caption{VersaGen can handle various forms of input provided by users, whether at the object-level, scene-level, or a combination of both.}
    \label{fig:fig1}
\end{figure}

\section{Introduction}

Ask any artist, and they will likely express their discontent with generative AI. They demand control over the creative process, control that aligns with their creative intent, and rightfully so! For you and me alike, beyond the initial "wow", questions are also starting to form around why models do not follow exactly what we want. There is a clear trend in the generative AI literature towards enhancing user control, which is encouraging. However, to fully understand the limitations of current approaches, it is essential to articulate the gap between human artistic expressivity and the capabilities of generative AI. To shed light on this, we first conducted a pilot study in which participants used Stable Diffusion (SD) \cite{rombach2022high} and ControlNet \cite{zhang2023adding} to generate images matching a given reference. The results revealed that participants faced two main challenges: incapability and inflexibility.

Incapability refers to the technical barriers that hinder users from achieving their desired visual outcomes, an issue widely acknowledged and tackled in recent works \cite{cao2023masactrl,ju2023humansd,ge2023expressive,gong2023check,wu2023harnessing,gafni2022make,ma2023subject,chang2023muse}. These solutions introduce various multi-modal input forms to make the process more intuitive. However, inflexibility remains largely underappreciated. Systems like ControlNet \cite{zhang2023adding} and T2I-Adapter \cite{mou2024t2i} require users to control the entire scene, which our findings suggest rarely facilitates meaningful engagement. In fact, 74.96\% of the 1,250 ControlNet trials reported that users found it exceedingly difficult to provide comprehensive visual guidance, particularly for complex scenes with multiple subjects.

To address the challenges of incapability and inflexibility, we propose a natural next step: allowing individuals the flexibility to decide what to control or not. Recognising that people express their creative intent differently -- we re-envisage the visual control problem in generative AI and transform it from a single visual subject/scene-level control to a hybrid approach comprising multiple degrees of control freedom.

Our proposed solution, VersaGen, targets drawing as the primary control tool, given its long-standing role in human expression of the visual world \cite{aubert2014pleistocene,karmiloff1990constraints,hoffmann2018u,gombrich1995story,jongejan2017quick,eitz2012humans}. Unlike previous systems that require compulsory holistic scene drawings, VersaGen admits four types of visual controls (Fig.~\ref{fig:fig1}): i) single visual subject; ii) multiple visual subjects; iii) scene background; iv) any combination of the three above or merely no control at all, falling back to pure text-to-image generation. By providing this flexibility and allowing users to choose the level of control that best suits their needs and preferences, VersaGen aims to make generative control more inclusive, accessible, and enjoyable for all, transforming the creative process into a fun and engaging experience. Achieving VersaGen is technically demanding, with two major challenges to overcome. The first challenge lies in simulating the various types of drawing inputs that VersaGen is designed to accommodate, given the scarcity of such data for training. To address this, we extract and optimise edge maps from images to imitate human drawing input. Although edges are not perfect substitutes for real drawings (\textit{e.g.}, human free-hand sketches), they provide a scalable way to obtain relevant training data. Additionally, we design an Adaptive Control Strength mechanism (Sec.~\ref{sec:in_the_wild}) to mitigate the disparity between edge maps and real drawings during inference. The key insight is that while drawing artifacts in realistic human input can degrade generation quality, their impact can be minimised by limiting their contribution to later steps in the diffusion chain.

The second challenge is visual localisation. Unlike existing works that require users to precisely position visual controls, VersaGen simplifies this process by automatically locating user-provided visual controls within the appropriate local context. To achieve this, we leverage recent advances showing that T2I models inherently function as semantic segmenters, capable of generating bounding boxes encircling objects based on their corresponding tokens at certain diffusion timesteps \cite{hertz2022prompt,zhang2021plug,patashnik2023localizing}. To further address potential localisation inaccuracies during inference, we introduce the Multimodal Conflict Resolver in VersaGen (Sec.~\ref{sec:MCR}). This component employs both token-level and pixel-level objectives to closely align the corresponding regions in the latent space across both modalities, thereby reducing conflicts and preventing unexpected generation results.

We conduct extensive evaluations using both edge maps and human free-hand sketches on the COCO \cite{lin2014microsoft} and Sketchy \cite{sangkloy2016sketchy} datasets. Our results demonstrate that VersaGen outperforms well-established T2I \cite{rombach2022high} and controllable T2I models \cite{mou2024t2i, zhang2023adding} in both quantitative and qualitative comparisons. Furthermore, a human study reveals that 48\% of users identify VersaGen as the most user-friendly interactive generation model compared to alternative approaches, underscoring the importance of providing flexible control options that cater to diverse user preferences and creative intents. Finally, a comprehensive ablation study highlights the crucial role of our three proposed strategies in enabling VersaGen to produce high-quality, user-controlled visual outputs across a wide range of input conditions in real-world scenarios.

\begin{figure*}[ht!]
    \centering
    \includegraphics[width=0.9\textwidth]{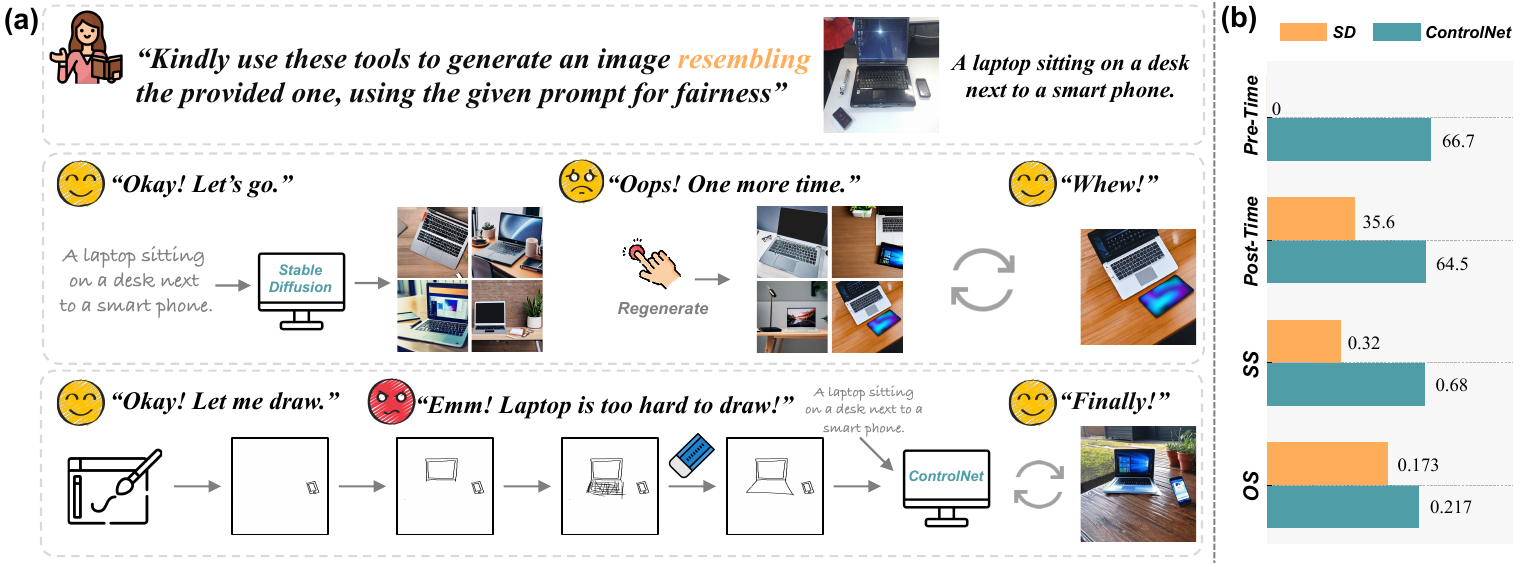}
    \caption{(a) The illustration of pilot study: users are tasked with generating an image similar to the given reference image using SD and ControlNet. (b) Quantitative evaluation of SD and ControlNet.}
    \label{fig:pilot}
\end{figure*}

\section{Related Work}
Recent advancements in diffusion models, such as Imegen \cite{saharia2022photorealistic}, Stable Diffusion (SD) \cite{rombach2022high}, and DALL-E \cite{ramesh2021zero}, have revolutionised text-to-image generation tasks. These models excel at producing high-quality images but often lack fine-grained control over the generated content. To address this limitation, researchers have explored various approaches to enhance user control in generative AI.
One line of research focuses on text-driven control methods, which involve adjusting prompts \cite{gal2022image,gani2023llm,kim2022diffusionclip,wang2023tokencompose,lee2023aligning,liu2022design,marcus2022very,wang2022diffusiondb} or improving cross-attention maps \cite{feng2022training,chefer2023attend,li2023divide,rassin2024linguistic,xu2024imagereward}. While these methods have shown promise in guiding the generation process, they often struggle to provide precise structural guidance, as text prompts alone may not fully capture the user's intended visual composition.
Another approach to controllable image generation involves incorporating additional input modalities, such as sketches or layouts. Layout Guidance Diffusion \cite{chen2024training}, GLIGEN \cite{li2023gligen} and InstanceDiffusion \cite{wang2024instancediffusion} leverage user-defined tokens and bounding boxes to guide cross-attention maps, while \cite{balaji2022ediff} imposes structural constraints on the generated images using similarity gradients between target sketches and intermediate model features. ControlNet \cite{zhang2023adding} and T2I-Adapter \cite{mou2024t2i} introduces a adapter to combine internal knowledge from text-to-image models with external control signals. UniControl \cite{qin2024unicontrol} builds on ControlNet and integrates task instructions into condition-specific networks to adapt to various visual inputs. Despite these advancements, existing methods often focus on single scene-level control, which can limit the degree of control available to users. In contrast, VersaGen takes a more flexible approach, accommodating both object-level and scene-level visual conditions without requiring the depiction of entire scene conditions. By offering a range of drawing options and allowing users to choose the level of control that best suits their needs, VersaGen empowers users with versatile control over the generative process.

\section{Pilot Study}
\label{sec:pilot}

To assess how well current off-the-shelf generative models meet user needs for generating target images, we designed this pilot study featuring Stable Diffusion \cite{rombach2022high}, a T2I model, and ControlNet \cite{zhang2023adding}, a controllable T2I model, as illustrated in Fig.~\ref{fig:pilot}(a). Users are provided with a reference image and a corresponding textual prompt, tasked with using SD and ControlNet to generate a matching image. In both experiments, participants are instructed to copy the prompt verbatim to maintain fairness. When using SD, users copy the given textual prompt, click ``Generate'', and then select a satisfactory generated image to ``Submit''. With ControlNet, additional preparation of visual conditions was required. Then, they copy the textual prompt, click ``Generate'', and wait for generated results, from which they select one to ``Submit''. Notably, i) a ``Regenerate'' button is provided, allowing users to regenerate images multiple times until they find a satisfactory result, at which point they can click ``Submit''. ii) Users cannot access the reference image after selecting the test generation model and must rely solely on their memory to complete the process. iii) Each time they click ``Submit'' during ControlNet trials, they are asked if providing an effective input condition for the current reference image is easy.

We employ four metrics to evaluate the two generative methods: i) \textbf{Pre-Time (s$\downarrow$):} The duration from when the user clicks the button to select a specific generative method to when they complete the preparation of input conditions and click the ``Generate'' button. This metric indicates the ease of use for end users. ii) \textbf{Post-Time (s$\downarrow$):} The time from when the method finishes generating a set of images to when the user clicks ``Submit''. Notably, users can regenerate results if they are unsatisfied with the current images. The Post-Time is accumulated, excluding the time spent waiting for model generation. iii) \textbf{Objective Similarity (OS$\downarrow$):} We use FID as a quantitative metric to evaluate the similarity between the given reference images and the user-submitted results. iv) \textbf{Subjective Similarity (SS$\uparrow$):} We invited five experts to evaluate the user-submitted results generated by the two methods. During each trial, the experts were presented with three images: reference, SD-generated, and ControlNet-generated. They were then asked to choose the one closer to the reference. The selection rate is referred to as SS.

\begin{figure*}[!htb]
    \centering
    \includegraphics[width=0.9\textwidth]{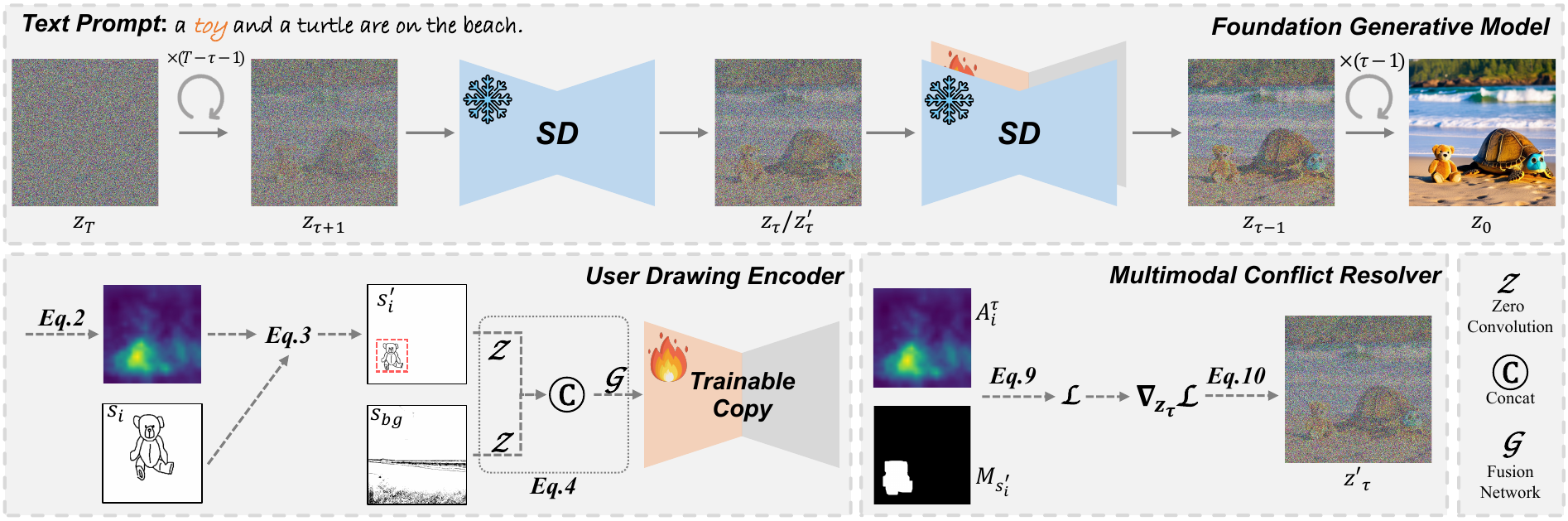}
    \caption{The illustration of VersaGen. At denoising timestep $\tau$ during inference, MCR functions to update the original noisy latent $z_\tau$ to $z^{'}_\tau$, to alleviate potential conflicts across modalities.}
    \label{fig:framework}
\end{figure*}

In Fig.~\ref{fig:pilot}(b), the findings show: i) There is no Pre-Time for SD as it solely relies on textual inputs, following instructions for participants to directly copy the given prompt. Conversely, considerable time is spent preparing visual conditions for ControlNet to enable more controlled generation. ii) Interestingly, participants spent more Post-Time on ControlNet than on SD. This indicates that for controllable generation models, good input conditions are crucial for achieving the desired results. It is not easy for participants to quickly find appropriate visual input conditions for ControlNet. iii) For the similarity evaluation, we observed that SD performs better in objective high-dimensional semantic feature measurement, as it is trained to generate semantically aligned images. However, for human perceptual visual similarity, ControlNet excels because the input visual conditions reflect the user's subjective interpretation of the reference image.

\section{Methods}
\textbf{Overview} 
The goal of VersaGen is to evolve human interaction with generative models from a strictly scene-level visual condition to a hybrid approach that supports multiple levels of drawing control. Users can freely select their preferred levels of visual conditions, eliminating the need for comprehensive scene-level conditions and bypassing the technical challenges associated with precise scene-level control. VersaGen accepts diverse drawing inputs: single visual subject drawing $s_i$, multiple visual subjects drawings $\{s_1, s_2, \cdots, s_m\}$, and background drawing $s_{bg}$. By synergising these elements with a simple textual prompt, users can realise highly flexible and controllable image generation. VersaGen comprises three core modules, as illustrated in Fig.~\ref{fig:framework}: i) \textbf{Foundation Generation Model (FGM)} incorporates the trainable copy provided by UDE to ``finetune'' its frozen weights, thereby aligning VersaGen's output with the user's drawings. ii) \textbf{User Drawing Encoder (UDE)} processes drawings inputted by users, encoding these hybrid drawings into a latent representation that serves as a condition to update a trainable copy of the Foundation Generative Model for fine-tuning. iii) \textbf{Multimodal Conflict Resolver (MCR)} addresses inconsistencies between modalities (user drawings and textual prompt) during inference, enhancing the quality of the generated images.

\subsection{Foundation Generative Model}
Here we adopt SD \cite{rombach2022high} as our FGM $\Phi$, a state-of-the-art text-to-image generation model. The process begins with an input image $z_0$, to which random noise is processively added, resulting in a noisy image $z_t$. With given textual condition $C$, a network $\epsilon_\theta$ is trained to predict the noise added to the noisy image $z_t$ with:
\begin{scriptsize}
\begin{equation}
    \mathcal L = \mathbb E_{z_t, t, C,  \epsilon \sim \mathcal N(0, 1)}[\Vert \epsilon - \epsilon_\theta(z_t, t, C)\Vert^2_2].
\end{equation}
\end{scriptsize}

Here $\mathcal L$ represents the overall learning objective of the entire diffusion model. Notably, the parameters in SD \cite{rombach2022high}, denoted as $\Theta_\Phi$, are kept fixed in the subsequent steps to preserve its foundational and powerful text-to-image generation capabilities.

\subsection{User Drawing Encoder}
Given visual subjects' drawings $S=\{s_1, s_2, \cdots, s_m\}$ and a textual prompt $C$, the initial step is to determine the potential locations of $s_i$ within the generated image $I$. We utilise the cross-attention mechanism in foundation generative model $\Phi$ to achieve this. Assuming the textual tokens correspond to $s_i$ are $c_i$, the cross-attention map $A^t_i$ is computed as follows, 
\begin{scriptsize}
\begin{equation}
\label{eq:crossattention}
    A^t_i(s_i, c_i) = \frac{1}{N}\sum\limits_{n=1}^N \mathrm {softmax}(\frac{Q^n(s_i)K^n(c_i)}{\sqrt{d}}).
\end{equation}
\end{scriptsize}

Here, $n$ refers to index of head in multi-heads mechanism, while $t$ denotes the timestep in diffusion process, notably, the $A^t_i$ is calculated only at $t=\tau$. $Q^n(s_i) \in \mathbb R^{B*HW*d}$ applies a linear function on $s_i$'s latent to generate ``query'' vector in cross-attention, where $H, W$ denote the height and width of latent, and $d$ is embedding dimension. Similarly, $K^n(c_i)\in \mathbb R^{B*d*77}$ involves a linear function followed by a transpose operation to produce ``key'' vector for cross-attention. 

To identify the potential location of $s_i$ within generated image $I(S, C)$, we first apply OTSU \cite{otsu1975threshold} on $A^t_i$ to compute the object-aware attention threshold, then extract potential mask $R_i$ of $s_i$, as Eq.\ref{eq: region} shown:
\begin{scriptsize}
\begin{equation}
\label{eq: region}
    \begin{aligned}
        R_i(x, y)  = 
        \begin{cases}
            1 & A^t_i(x,y) \geq \mathrm{OTSU}(A^t_i) \\
            0 & \mathrm{otherwise}
        \end{cases} ,
    \end{aligned}
\end{equation}
\end{scriptsize}
where $x$ and $y$ are coordinates of the element in $R_i$. Subsequently, the square bounding box $B_i$ of non-zero region in $R_i$ is used to guide the relocation of $s_i$, resulting in $s^{'}_i$, which is input into drawing encoder $\mathcal Z$ parameterised by $\theta_{z1}$. The latents of user drawings $S$ are concatenated, and then passed through a small fusion network $\mathcal G$ to produce a combined drawing latent $z_S$, as shown in Eq.\ref{eq:zs}:
\begin{scriptsize}
\begin{equation}
\begin{aligned}    
\label{eq:zs}
        z_S = \mathcal G(\mathrm{Concat}[ \mathcal Z(s^{'}_1;\theta_{z1}), \mathcal Z(s^{'}_2;\theta_{z1}), \\
        \cdots, \mathcal Z(s^{'}_m;\theta_{z1}), \mathcal Z(s_{bg};\theta_{z1})]; \theta_g).\\
\end{aligned}
\end{equation}
\end{scriptsize}

To leverage the T2I capabilities of $\Phi$, we adopt the ControlNet \cite{zhang2023adding} pipeline to add spatial control capability to $\Phi$. ControlNet \cite{zhang2023adding} creates a trainable copy of UNet in $\Phi$, which features zero convolution layers within the encoder blocks and the middle block. This process is formulated as:
\begin{scriptsize}
\begin{equation}
\label{eq:ys}
    y_S = \underbrace{\mathcal F(z_t; \Theta_\Phi)}_{\mathrm {FGM}} + \underbrace{\mathcal Z(\mathcal F(z_t+z_S; \Theta_S); \theta_{z2})}_{\mathrm{UDE}},
\end{equation}
\end{scriptsize}
where $\mathcal F$ is UNet, $\Theta_\Phi$ is the frozen weight of pretrained $\Phi$, $\mathcal Z$ corresponds to zero convolutions with learnable weights $\theta_{z2}$, $\Theta_S$ is the parameters of trainable copy of UNet. The whole training object of VersaGen is formulated as:
\begin{scriptsize}
\begin{equation}
    \mathcal L = \mathbb E_{z_t, t, C, z_S,  \epsilon \sim \mathcal N(0, 1)}[\Vert \epsilon - \epsilon_\theta(z_t, t, C, z_S)\Vert^2_2].
\end{equation}
\end{scriptsize}
\begin{table*}[!t]
    \setlength{\tabcolsep}{1mm}
    \small
    \centering
    \begin{tabular}{lcc|cccc|cccc}
    \toprule
    \multirow{2}{*}{Methods} & \multicolumn{2}{c}{w/GT} & \multicolumn{4}{c}{COCO} & \multicolumn{4}{c}{Sketchy} \\
    
            \cdashline{2-11}
    		& LoD & BBoxes & CLIP $\uparrow$ & FID $\downarrow$ & DINO $\uparrow$ & ACC $\uparrow$ & CLIP $\uparrow$  & FID $\downarrow$ & DINO $\uparrow$ & ACC $\uparrow$ \\
    		\cline{1-11}
    		Stable Diffusion \cite{rombach2022high} & / & / & \textbf{0.2556}  & 0.4026 & 0.7616 & 0.6803
     & 0.2509  & 0.3064 & 0.7934 & 0.7710 \\
            % \cdashline{1-11}
    		T2I-Adapter \cite{mou2024t2i} & $\checkmark$ & $\times$ & 0.2439 & 0.3732 & 0.7722
     & 0.6143 & 0.2392  & 0.3290 & 0.7843 & 0.7182 \\
            ControlNet \cite{zhang2023adding} & $\checkmark$ & $\times$ & 0.2523 & 0.3560 & 0.7774
     & 0.6547 & 0.2517  & 0.3493 & 0.8007 & 0.7622 \\
            UniControl \cite{qin2024unicontrol} & $\checkmark$ & $\times$ & 0.2467 & 0.3558 & 0.7475 & 0.7006 & 0.2479 
    & 0.3165 & 0.7988 & 0.8096 \\
            % \cdashline{1-11}
            GLIGEN \cite{li2023gligen} & $\times$ & $\checkmark$ & 0.2515 & 0.3571 & 0.7740 & 0.4319 
    & 0.2419 & 0.2961 & 0.8149 & 0.5550 \\
            InstanceDiffusion \cite{wang2024instancediffusion}  & $\times$ & $\checkmark$ & 0.2419 & 0.3811 & 0.7644 & \textbf{0.8094}
    & 0.2399 & 0.2818 & 0.8179 & \textbf{0.8942} \\
            % \cdashline{1-11}
    		VersaGen (Ours) & $\checkmark$ & $\times$ & 0.2537  & \textbf{0.3377} & \textbf{0.8019}
     & 0.7322 & \textbf{0.2524} & \textbf{0.2614} & \textbf{0.8193} & 0.8328\\
            VersaGen (Ours) & $\times$ & $\times$ & 0.2542  & \textbf{0.3431} & \textbf{0.7879} & 0.7229 & \textbf{0.2542} & 0.3080 & 0.8088 & 0.8174\\
    		\bottomrule
    	\end{tabular}
     \caption{Quantitative results. ``w/ GT'' denotes the locations of drawings (LoD) or bounding boxes (BBoxes) from GT.}
    \label{tab:comparison} 
\end{table*}

\begin{figure*}[ht!]
    \centering
    \includegraphics[width=0.9\textwidth]{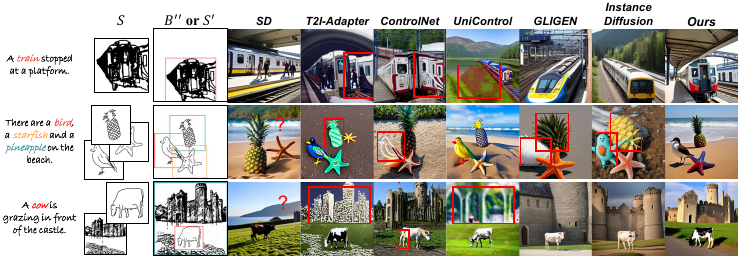}
    \caption{Visualised comparison of SD, T2I-Adapter, ControlNet, UniControl, GLIGEN,  InstanceDiffusion and our proposed VersaGen. Problematic regions are highlighted with $\color{red}\Box$, and missing entities are indicated by $\color{red}?$.}
    \label{fig:Comparsion}
\end{figure*}
\subsection{Multimodal Conflict Resolver} 
\label{sec:MCR}
For multimodal controllable image generation, generating a high quality realistic image hinges on ensuring that the latent can coherently integrate conditional information from different modalities. During the training phase of VersaGen, the textual prompt $C$ is carefully aligned with the drawings $S$ because both are derived from the ground truth image $z_0$. However, during inference, we cannot guarantee that the user-provided drawings are always coherent with the accompanying textual prompt which may result in a modality conflict during the generation process. Therefore, we introduce a Multimodal Conflict Resolver (MCR) during inference to mitigate the impact of such issues.

Inspired by \cite{wang2023tokencompose}, we address this issue by a test-time optimisation. Our objective is to ensure that the active region in the cross-attention map $A^t_i(z_t, c_i)$ closely aligns with the contour $M_{s^{'}_i}$ of $s^{'}_i$. Therefore we involve a token-level loss and a pixel-level loss to achieve this goal, as formulated by:
\begin{scriptsize}
\begin{equation}
    \mathcal{L}_{token} = \frac{1}{m}\sum\limits_{l=1}^{L} \sum\limits_{i=1}^{m}\left(1-\frac{\sum\limits_{x=1}^{h_l}\sum\limits_{y=1}^{w_l}{M_{s^{'}_i}(x, y)A_{il}^t(x, y)}}{\sum\limits_{x=1}^{h_l}\sum\limits_{y=1}^{w_l}{A_{il}^t(x, y)}}\right)^2,
\end{equation}
\end{scriptsize}

\begin{scriptsize}
\begin{equation}
    \mathcal L_{pixel}  = \frac{1}{m}\sum\limits_{l=1}^L\sum\limits_{i=1}^m \sum\limits_{x=1}^{h_l}\sum\limits_{y=1}^{w_l}\left(\mathrm{BCE}(M_{s^{'}_i}(x, y), A^t_{il}(x, y))\right),
\end{equation}
\end{scriptsize}

\begin{scriptsize}
\begin{equation}
    \mathcal{L} = \lambda \mathcal{L}_{token} + (1-\lambda)\mathcal{L}_{pixel},
\end{equation}
\end{scriptsize}
where $l$ is the layer index of the UNet, $\mathrm{BCE}(\cdot, \cdot)$ represents binary cross entropy loss. Since different layers have different resolutions of $A^t_{il}$, we downscale the resolution of $M_{s^{'}_i}$ to match that of $A^t_{il}$ using bilinear interpolation, followed by binarisation of all values. Then we using $\mathcal{L}$ to update $z_t$:
\begin{scriptsize}
\begin{equation}
    z^{'}_t = z_t - \alpha \nabla_{z_t}\mathcal{L},
\end{equation}
\end{scriptsize}
where $\alpha$ is a hyperparameter used to adjust the gradient update step size. It is important to note that MCR only optimises $z_t$ at $t=\tau$, resolving the modality conflict at the timestep immediately following the incorporation of user drawing control. Once the optimisation is completed, MCR ceases operation in subsequent timesteps to minimise resource use.
\subsection{VersaGen In-the-Wild}
\label{sec:in_the_wild}
In real-world applications of VersaGen, two main challenges arise: i) when users provide multiple visual subjects drawings, the correlation between objects causes significant overlap in the object-aware active region $R_i$s, leading to low-quality, occluded outputs. ii) Amateur users often struggle to create precise, high-quality drawings, resulting in unrealistic generated images. To improve VersaGen's real-world adaptability, we propose two strategies during inference: multi-object decoupling and adaptive control strength.
\begin{figure*}[t]
    \centering
    \includegraphics[width=0.9\textwidth]{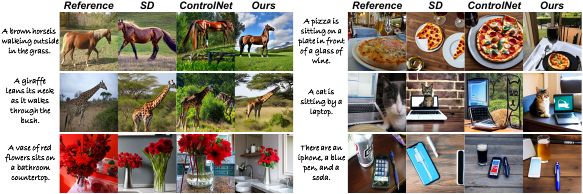}
    \caption{Visualisation of the reference images used in the human study alongside the user-submitted results generated by the three methods.}
    \label{fig:human}
\end{figure*}

\textbf{Multi-object Decoupling.} Given the user drawings $S=\{s_1, s_2, \cdots, s_m\}$, their corresponding active regions $R_i$ in cross-attention map can be calculated by Eq.\ref{eq:crossattention}. Each $R_i$ can be simplified as a bounding box $B_i = (x^{min}_i, y^{min}_i, x^{max}_i, y^{max}_i)$, where $(x^{min}_i, y^{min}_i)$ are the coordinates of the top-left corner of $B_i$, $(x^{max}_i, y^{max}_i)$ are bottom-right corner. The goal of multi-object decoupling is to ensure the Intersection over Union (IoU) of $B_i$s remains within a reasonable range, avoiding excessive overlapping. The centre point of each $B_i$ is given by $(c^{i}_x, c^{i}_y) = (\frac{x^{max}_i-x^{min}_i}{2}, \frac{y^{max}_i-y^{min}_i}{2})$. The overall centre of all $B_i$s is calculated as $(c_x, c_y)=(\frac{\sum_{i=1}^m c^i_x}{m}, \frac{\sum_{i=1}^m c^i_y}{m})$. For any two visual subjects drawings $s_i$ and $s_j$, we calculate the IoU between $B_i$ and $B_j$. If $\mathrm{IoU}(s_i, s_j) \geq \beta$, we apply two operations on both $B_i$ and $B_j$ simultaneously: i) \textit{resizing}, ii) \textit{translation}, to adjust their bounding boxes and reduce overlap.  

\textit{Resizing} shrinks the region of $B_i$ by two units towards $(c^{i}_x, c^{i}_y)$,
\begin{scriptsize}
\begin{equation}
    B^{'}_i = (x^{min}_i+1, y^{min}_i+1, x^{max}_i-1, y^{max}_i-1).
\end{equation}
\end{scriptsize}

\textit{Translation} moves $B_i$ by one unit away from $(c_x, c_y)$. First, we calculate the direction vector $\Vec{d} = (c_x - c^i_x, c_y - c^i_y)$, and then obtain the unit direction vector $\hat d$. We then translate $B^{'}_i$ to $B^{''}_i=(x^{''min}_i, y^{''min}_i, x^{''max}_i, y^{''max}_i)$ as follows:
\begin{scriptsize}
\begin{equation}
    \begin{aligned}
        (x^{''min}_i, y^{''min}_i) &= (x^{'min}_i, y^{'min}_i) - \hat d,\\
        (x^{''max}_i, y^{''max}_i) &= (x^{'max}_i, y^{'max}_i) - \hat d. \\
    \end{aligned}
\end{equation}
\end{scriptsize}

In practice, we apply these two operations in combination until $\mathrm{IoU}(s_i, s_j) < \beta$. More details about the combination strategy can be found in the supplemental material\cite{chen2024versagen}.

\textbf{Adaptive Control Strength.} For amateur end users, the goal of providing drawing control is not to make the generated image strictly follow their strokes, as they understand their limited drawing skills may not fully represent their intentions. This makes the textual prompt crucial for achieving their desired results. To address this, we design an adaptive control strength strategy that reduces the influence of user drawings as the denoising timesteps progress. In the early structure-forming stage, user drawings dominate to establish the basic structure. In the later detail generation stage, the textual prompt takes over, adding coherent details based on its well-pretrained text-to-image (T2I) capabilities. This strategy is implemented by introducing a weight $\gamma(t)$ to Eq.\ref{eq:ys}:
\begin{scriptsize}
\begin{equation}
\begin{aligned}
    y_S &= \mathcal F(z_t; \Theta_\Phi) + \gamma(t)\mathcal Z(\mathcal F(z_t+z_S; \Theta_S); \theta_{z2}), \\
    \gamma(t) &= 1-a\cdot\frac{1}{1+e^{-b\cdot\left(t-c\right)}},
\end{aligned}
\end{equation}
\end{scriptsize}
where $a, b, c$ are used for timestep-aware adjustment. 
\section{Experiments}
VersaGen utilises SD as Foundation Generation Model. We conducted training and testing of VersaGen on COCO \cite{lin2014microsoft} and further assessed its performance on Sketchy \cite{sangkloy2016sketchy}, a human free-hand sketch dataset. 
Detailed information about data processing, evaluation metrics hyperparameters employed in the experiments, and additional generated results are provided in the supplementary\cite{chen2024versagen}.

%%% 图的格式要设置
\begin{figure*}[h]
    \centering
    \includegraphics[width=0.9\textwidth]{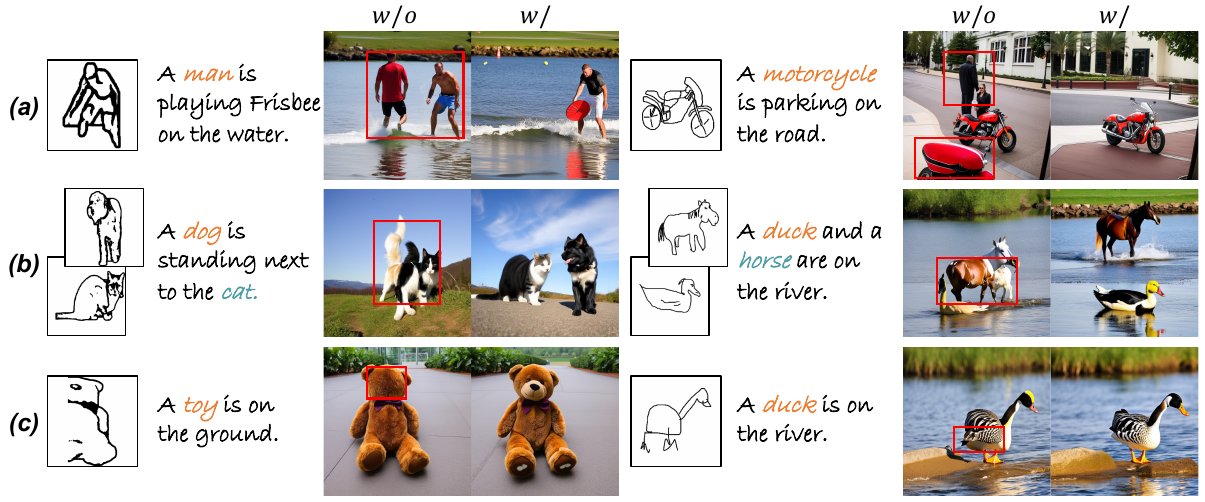}
    \caption{Ablative study of different designs in VersaGen: (a) Multimodal Conflict Resolver, (b) Multi-object Decoupling, (c) Adaptive Control Strength. Problematic regions are highlighted with $\color{red}\Box$.}
    \label{fig:ablation}
\end{figure*}
\subsection{Qualitative Results}
\textbf{\textit{Main results of VersaGen.}} 
Fig.~\ref{fig:fig1} shows our results with various drawing inputs: \textit{single visual subject}, \textit{multiple visual subjects} and \textit{subjects+background}. The images generated by VersaGen not only closely align with the text descriptions but also maintain excellent consistency with the appearance of user drawings. VersaGen can automatically place objects from separate drawings coherently within a background, guided by the textual prompt.

\textbf{\textit{Comparison with different methods.}}
Fig.~\ref{fig:Comparsion} presents qualitative results corresponding to two types of drawings (Row$1$ for Edge Map, Row$2$ \& $3$ for Free-Hand Drawing). As these comparative methods are unable to process object-level drawings, we ensure a fair comparison by using the automatically identified bounding box $B^{''}$ or relocated user drawings $S^{'}$ from VersaGen as their visual conditions. We can observe that: 
i) From column $S$ to $B^{''}$ or $S^{'}$, VersaGen effectively relocates the $s_i$s to positions that align coherently with the given textual prompt. Moreover, even in scenarios involving multiple visual subjects, VersaGen successfully avoids overlapping locations. 
ii) For the SD results, when the textual prompt involves multiple visual subjects, some entities are frequently omitted, \textit{e.g.}, The ``bird'' in Row$2$ and the ``castle'' in Row$3$ are missing. 
iii) The results from T2I-Adapter, ControlNet and UniControl adhere too closely to the provided drawings, often resulting in the generation of unrealistic details, such as the ``bird'' in Row$2$. 
iv) The results from GLIGEN and InstanceDiffusion indicate that employing layouts $B^{''}$ as a condition does not yield precise object control. The presence of a large number of irrelevant areas in the layout diminishes the positive influence of MOD on reducing object overlap, which leads to a decline in the quality of object generation, \textit{e.g.}, the ``bird'' and ``pineapple'' in Row$2$.
v) Our proposed VersaGen consistently outperforms other methods by generating high-quality images with precise semantics across a range of user drawing combinations, such as single visual subject, multiple visual subjects, and subjects+background.
\subsection{Quantitative Results}

Tab.~\ref{tab:comparison} shows quantitative results on COCO and Sketchy. 

For COCO results, we can see that: i) Compared to SD, other methods achieved lower FID scores due to the additional drawings or layouts, enhancing the spatial alignment between the generated images and the ground truth.
ii) Our VersaGen achieves better FID metrics regardless of whether ground truth locations or automatically predicted locations are used, demonstrating the robustness and superiority of VersaGen. 
iii) SD achieves the highest CLIP score, likely due to the conflict of multiple conditions or the limitations in edge map extraction techniques. VersaGen addresses this by substituting the latter with high-quality user drawings, indicating the effectiveness of our proposed strategies during inference.
iv) VersaGen achieves the highest DINO score with different setting, which indicates the generated images are closest to the real images in visual features, further demonstrating the utility of our method. 
v) InstanceDiffusion excels in object semantics, achieving the highest ACC but at the cost of local object details. In contrast, using sketches as conditions better balances local details with global semantics.

For Sketchy, visual subject drawings are human free-hand sketches, which are less precise and more abstract than edge maps, reflecting real-world conditions where most users are amateurs and may struggle to produce high-quality drawings quickly. Despite these challenges, our method achieves the highest performance on CLIP score, FID, and DINO score on Sketchy, demonstrating VersaGen's effectiveness in real-world scenarios.

\subsection{User Study}
\label{sec:UserStudy}
To further assess VersaGen's effectiveness, we conducted a user study with the same setup as described in Sec.\ref{sec:pilot}. The study presented user-submitted results from three methods (Fig.~\ref{fig:human}), revealing that: i) all methods perform well for simple prompts with a single entity; ii) however, for complex prompts with multiple entities, both SD and ControlNet often miss entities, and strict adherence to input conditions by ControlNet can degrade output quality. We surveyed 50 participants, asking which method balances ease of use with reliable generation capability. Results show 48\% favor VersaGen, underscoring its practical application potential.

\subsection{Ablation Study}

To enhance the practicality of VersaGen in real-world applications, we propose three inference-phase strategies: Multimodal Conflict Resolver (MCR), Multi-object Decoupling (MoD), and Adaptive Control Strength (ACS). We conduct an ablation study to thoroughly evaluate the impact and effectiveness of these strategies. i) As shown in Fig.~\ref{fig:ablation}(a), the absence of MCR results in inconsistencies between the textual prompt and the generated images; for instance, the prompt ``A man'' yields an image of two men. ii) Without MoD, overlapping multiple objects leads to poor-quality generations, as depicted in Fig.~\ref{fig:ablation}(b), such as a creature with a horse's body and a duck's head. Involving MoD effectively resolves this issue. iii) Fig.~\ref{fig:ablation}(c) demonstrates that with ACS, even user-provided low-quality drawings, VersaGen can also generate high-quality images with details.

\section{Conclusion} 
In this paper, we present VersaGen, a generative AI agent that addresses the challenges of incapability and inflexibility in text-to-image synthesis by providing users with versatile visual control options. By allowing users to choose the level of control that best suits their needs and preferences, VersaGen transforms the creative experience into a more engaging and fulfilling one. We hope our work inspires further research in developing user-centric generative AI solutions that prioritise flexibility, accessibility, and are ultimately people-facing.

\section{Acknowledgments}
This research was supported by the Science and Technology Innovation 2030 – Major Projects (Grant No. 2021ZD0200600), the Hainan Provincial Natural Science Foundation of China (Grant No. 624LALH008), and the Program for Youth Innovative Research Team of BUPT (Grant No. 2023YQTD02). The authors sincerely appreciate the support and opportunities provided by these projects, which were instrumental in the successful completion of this research.

\bibliography{aaai25}

\clearpage
\appendix

\section{Supplemental material}
This document includes implementation parameters, experimental details and additional results for the main paper.

\subsection{Implementation details}
\subsubsection{Multi-object Decoupling}

During Multi-object Decoupling process, after each \textit{resizing} operation, we perform multiple \textit{translation} attempts to ensure that $\mathrm{IoU}(s_i, s_j) < \beta$. We make a maximum of 10 \textit{translation} attempts to achieve this goal. The entire iterative process is repeated at most 25 times. We set $\beta$ to 0.1.

\subsubsection{Evaluation Metrics}
We evaluated our method and comparison
methods using four metrics: CLIP score \cite{radford2021learning} evaluates the semantic consistency between text and images; FID \cite{heusel2017gans} and DINO \cite{oquab2023dinov2} measure the perceptual similarity between generated and original images. Additionally, we used Grounding DINO \cite{liu2023grounding} to detect entities in generated images, considering a match with the text as a correct sample. Object accuracy is the proportion of these correct samples. We set the box threshold to 0.3 for Ground DINO object detection.

\subsubsection{Data Preprocessing}

\textbf{COCO \cite{lin2014microsoft} } Due to the lack of hand-drawn sketches in the COCO, we adopted a strategy of extracting the edge maps of objects in the images as pseudo sketches. The steps for this process are as follows:

i) Use SAM \cite{kirillov2023segment} to perform precise segmentation of objects within the images, obtaining a series of entity segmentation maps $\{Mask_1, Mask_2, ..., Mask_n\}$. 

ii) Utilise the Spacy \cite{honnibal2020spacy} natural language processing tool to extract object nouns $\{c_1,c_2,...,c_m\}$ from the image captions and record the positional indices $\{index_{c_1},index_{c_2},...,index_{c_m}\}$ of these nouns in the captions. 

iii) Perform target detection on the images with the GroundingDINO \cite{liu2023grounding} model, and then filter out the targets corresponding to the extracted nouns $\{c_1,c_2,...,c_m\}$.

iv) Use the segmentation map $\{{Mask}_1, Mask_2, ..., Mask_n\}$ to precisely segment the target objects from the background, ensuring that the separated objects have a uniform solid colour background.

v) Apply pseudo graffiti techniques \cite{zhang2023adding} to extract edge maps from object images with a solid colour background, generating pseudo sketches $\{s_1, s_2,...,s_k\}$.

Following the aforementioned data processing, we have compiled a dataset comprising 285,000 training and 3,000 testing triplets, each consisting of [sketches, image, caption]. 

\textbf{Sketchy \cite{sangkloy2016sketchy}}
The Sketchy, while comprehensive in its provision of category labels, lacks the accompaniment of captions for the images it contains. To address this limitation, we have employed the BLIP \cite{li2022blip} model to generate descriptive captions for each individual image within the database. Through this meticulous data processing endeavour, we have successfully curated a collection of 5,000 triplets ({sketch, image, caption}) and are thus suitable for testing and further analysis.

\textbf{Training and Inference Parameters}
Our model is based on  Stable Diffusion V2.1. In the training phase, we employ a regimen of data augmentation to enhance the model’s generalisation capabilities. This involves the random omission or inclusion of lines from the input sketch conditions. In order to focus the model’s attention on the primary objective of conditional image generation, we utilise the GT locations directly.

The training process is conducted over a total of 10 epochs, with a learning rate of ${10}^{-5}$, and the optimiser is AdamW \cite{loshchilov2018decoupled}. It was trained for 12 hours on 3 NVIDIA Tesla A800 80G GPUs, with a batch size of 36. For the denoising process, we set the number of steps, $T$, to 50, and the denoising time threshold, $\tau$, is adjusted to 48.We set the hyperparameters $a,b,c,\alpha,\beta$ and $\lambda$ to $0.7,0.6,15,15,0.1$ and $0.05$,respectively.

\subsection{Pilot study and user study}
\subsubsection{Experimental Design and Result Comparison} 

Note: This study does not involve any potential risks to personal safety and has obtained approval from the relevant institution.

We randomly selected 50 images from COCO as reference images in the study. Each user is asked to generate an image based on a reference image using a variety of image generation methods (2 methods for the pilot study in sec.\ref{sec:pilot}, 3 methods for user study in sec.\ref{sec:UserStudy}). To motivate the users, we have invited five experienced experts to evaluate these generated images. The more similar the generated images are to the original reference images, the higher the reward the users will receive. We organise 25 users aged from 14 to 55 to participate in this experiment, coming from various professional backgrounds, including high school students, college students, food delivery workers, programmers, baristas, chefs, and waitstaff, to ensure that we can collect a wide and diverse range of feedback. 

The results of Our VersaGen in user study are as follows:

i) Pre-Time is 48.6 and Post-Time is 58.7, they are both less than ControlNet (48.6 \textit{v.s.} 66.7, 58.7 \textit{v.s.} 64.5). 

ii) SS is 0.51, The sum of the SS scores of the three methods is 1, at which point the SS of SD and ControlNet are 0.19, and 0.30 respectively.

iii) OS is 0.163, it better than SD and ControlNet (0.163 \textit{v.s.} 0.173 \textit{v.s.} 0.217).

\subsubsection{How VersaGen Reduces Pre/Post-Time}
Based on the design of the user study in sec.\ref{sec:UserStudy}), we conducted an extended experiment to explore how VersaGen reduces the pre-time and post-time of image generation from user feedback. We conducted a follow-up analysis based on our user study where 150 participants performed image generation tasks using VersaGen and ControlNet, followed by a survey to gather feedback on the time-saving aspects of each method. The survey questions focused on identifying the key factors contributing to the reduction in pre-time and post-time when using VersaGen.

Participants were asked to select reasons for pre-time and post-time reduction from the following options:
\begin{itemize}
    \item \textbf{Pre-time Reduction}: Faster condition preparation (FCP), Simpler target images (STI), Greater proficiency in operation (GPO).
    \item \textbf{Post-time Reduction}: Fewer generation attempts (FGA), Greater proficiency in operation (GPO), Lack of interest in further participation (LIP).
\end{itemize}

The comparison results are shown in Tab.~\ref{tab:UserStudyEx}.
\begin{table}[htb]
    % \small
    \centering
    \setlength{\tabcolsep}{0.7mm}
    \begin{tabular}{lccccc}
    \toprule
    Reason & FCP (\%) & STI (\%) & GPO (\%) & FGA (\%) & LIP (\%) \\
    \cline{1-6}
    Pre-Time & 69 & 21 & 10 & - & -\\
    Post-Time & - & - & 14 & 77 & 9\\
    \bottomrule
    \end{tabular}
     \caption{The statistics of the reasons for how VersaGen reduces the pre/post-time.}
    \label{tab:UserStudyEx} 
\end{table}

The data of Tab.~\ref{tab:UserStudyEx} indicates that the primary factor contributing to reduced pre-time was faster condition preparation (FCP). VersaGen's support for user-drawn input significantly streamlines the initial setup compared to ControlNet, making it more user-friendly and efficient for preparing inputs.

For post-time reduction, most users indicated that fewer generations were attempted (FGA) to achieve satisfactory results, suggesting that VersaGen's enhanced generation capabilities better align with users' creative intentions and produce higher-quality images. This minimizes the need for multiple iterations and refinements, leading to faster task completion.

Overall, the results highlight that VersaGen's intuitive input methods and robust generation capabilities improve both pre- and post-time efficiency, enhancing the overall user experience.

\subsubsection{Satisfaction from Adding Drawings}
We conducted a follow-up survey involving 150 participants in image generation tasks using VersaGen, SD, and ControlNet to assess overall satisfaction with the use of drawing inputs. The evaluation focused on two main aspects:
\begin{itemize}
    \item \textbf{Satisfaction with Generated Images}: Far from the target (FFT), Similar but poor quality (SPQ), Partially similar with flaws (PSF), Close with flaws (CWF), Very close to the target (VCT).
    \item \textbf{Acceptance of Drawing Input}: Drawings suit intention (DSI), Interested in drawing (ID), Good at drawing (GAD); Drawings are troublesome (DIT), Drawings are inconvenient (DI), Hard to control strokes (HCS).
\end{itemize}

The statistical results are shown in Tab.~\ref{tab:Satisfaction} and Tab.~\ref{tab:Acceptance}.
\begin{table}[htb]
    \small
    \centering
    \setlength{\tabcolsep}{0.6mm}
    \begin{tabular}{lccccc}
    \toprule
    Methods & FFT(\%) & SPQ(\%) & PSF(\%) & CWF(\%) & VCT(\%) \\
    \cline{1-6}
    Stable Diffusion & 35 & 48 & 12 & 4 & 1\\
    ControlNet & 22 & 36 & 21 & 18 & 3\\
    ours & 5 & 17 & 33 & 35 & 10\\
    \bottomrule
    \end{tabular}
     \caption{The statistics of user satisfaction with the generated images}
    \label{tab:Satisfaction} 
\end{table}

i) The results of Tab.~\ref{tab:Satisfaction} indicate that VersaGen produces images closest to the target, with the highest percentage of "Close with flaws" (CWF) and "Very close to the target" (VCT) ratings, followed by ControlNet and SD.

\begin{table}[htb]
    % \small
    \centering
    \setlength{\tabcolsep}{1.3mm}
    \begin{tabular}{lccc}
    \toprule
    DSI (\%) & ID (\%) & GAD (\%) & Total Accept (\%) \\
    \cdashline{1-4}
    46 & 17 & 6 & 69\\
    \cline{1-4}
    DIT (\%) & DI (\%) & HCS (\%) & Total Reject (\%) \\
    \cdashline{1-4}
    15 & 12 & 4 & 31\\
    \bottomrule
    \end{tabular}
     \caption{The statistics of user acceptance of Drawing as a conditional input.}
    \label{tab:Acceptance} 
\end{table}

ii) These insights of Tab.~\ref{tab:Acceptance} underline that the use of drawing inputs in VersaGen is well-received by a majority of users, providing a significant advantage in aligning outputs with the intention of users.

\subsection{Exploration of Generalization}
\subsubsection{The Influence of Hyperparameters} When adapting our method to other models (e.g., SDXL\cite{podellsdxl}), only two MCR hyperparameters— $\alpha$ (gradient step size) and $\lambda$ (balance between token and pixel loss)—need tuning for high-performance requirements. However, extensive hyperparameter adjustments are generally unnecessary. We used the same hyperparameter configuration from the experiment of Tab.~\ref{tab:comparison} on a different T2I model (from SD2.1 to SDXL). The results of Tab.~\ref{tab:hyperparameters} demonstrated substantial improvements across all metrics. The experimental dataset is COCO.
\begin{table}[htb]
    % \small
    \centering
    \setlength{\tabcolsep}{1.3mm}
    \begin{tabular}{lccccc}
    \toprule
    Methods & w/GT & CLIP $\uparrow$ & FID $\downarrow$ & DINO $\uparrow$ & ACC $\uparrow$ \\
    \cline{1-6}
    SDXL & / & 0.2569 & 0.4152 & 0.7757 & 0.6428\\
    ControlNet & $\checkmark$ & 0.2558 & 0.3361 & 0.8031 & 0.7075\\
    Ours & $\times$ & \textbf{0.2596} & \textbf{0.3110} & \textbf{0.8158} & \textbf{0.7943}\\
    \bottomrule
    \end{tabular}
     \caption{The comparison results of the same hyperparameters on model generalization.}
    \label{tab:hyperparameters} 
\end{table}

\subsubsection{The Versatility of MCR} To demonstrate the versatility of MCR, we integrated it with the trained ControlNet and compared its performance. The experimental results are shown in Tab.~\ref{tab:versatility}. The fundamental model of the used ControlNet is SDXL. The experimental dataset is COCO.
\begin{table}[htb]
    % \small
    \centering
    \setlength{\tabcolsep}{0.7mm}
    \begin{tabular}{lccccc}
    \toprule
    Methods & w/GT & CLIP $\uparrow$ & FID $\downarrow$ & DINO $\uparrow$ & ACC $\uparrow$ \\
    \cline{1-6}
    ControlNet & $\checkmark$ & 0.2558 & 0.3361 & 0.8031 & 0.7075\\
    ControlNet + MCR & $\checkmark$ & \textbf{0.2562} & \textbf{0.3303} & \textbf{0.8067} & \textbf{0.7284}\\
    \bottomrule
    \end{tabular}
     \caption{The effectiveness comparison of the combination of ControlNet and MCR.}
    \label{tab:versatility} 
\end{table}

\subsection{Ablation Study of Different Levels Visual Control}
We conducted an additional ablation study focusing on the different levels of visual control using the same setup as Tab.~\ref{tab:comparison} and the COCO dataset for consistency. We processed 1,000 test samples by selecting images with multiple visual subjects and isolating scene backgrounds using segmentation masks.

For each processed test sample, we evaluated four distinct scenarios:
\begin{itemize}
    \item Text-only (no visual control)
    \item Single visual subject control
    \item Multiple visual subjects control
    \item Multiple visual subjects with a scene background
\end{itemize}
\begin{table}[htb]
    \small
    \centering
    \setlength{\tabcolsep}{0.7mm}
    \begin{tabular}{lccccc}
    \toprule
    Control Type & w/GT & CLIP $\uparrow$ & FID $\downarrow$ & DINO $\uparrow$ & ACC $\uparrow$ \\
    \cline{1-6}
    Text-only (no control) & $\times$ & 0.2508 & 0.4948 & 0.7746 & 0.6259\\
    Single Visual Subject & $\times$ & 0.2557 & 0.4249 & 0.7806 & 0.6590\\
    Multiple Visual Subjects & $\times$ & 0.2572 & 0.4168 & 0.7902 & 0.7182\\
    Mult. Vis. Subj. + Sc. Bg. & $\times$ & \textbf{0.2580} & \textbf{0.4061} & \textbf{0.8103} & \textbf{0.7264}\\
    \bottomrule
    \end{tabular}
     \caption{The ablation experiment results of visual control at different levels. For convenience of presentation, "Mult. Vis. Subj. + Sc. Bg." represents "Multiple Visual Subjects + Scene Background."}
    \label{tab:DifferentLevels} 
\end{table}

It is known from the experimental results from Tab.~\ref{tab:DifferentLevels} that: 

i) The results show a clear trend: as the level of visual control increases, performance metrics improve. Starting from text-only input (which had the lowest scores), adding a single visual subject enhanced the results, and incorporating multiple visual subjects provided further gains. This indicates that richer visual cues contribute significantly to better image generation. 

ii) After the incorporation of scene background information, all metrics have witnessed improvement. This further demonstrates the robust generalization of VersaGen across diverse control levels. Moreover, with the addition of more original image information, the generated images exhibit higher similarity to the originals.

These findings confirm the benefits of incorporating different levels of visual control and provide a deeper understanding of how scene context impacts image generation.

\subsection{Additional results}
\begin{figure*}[!h]
    \centering
    \includegraphics[width=0.98\textwidth]{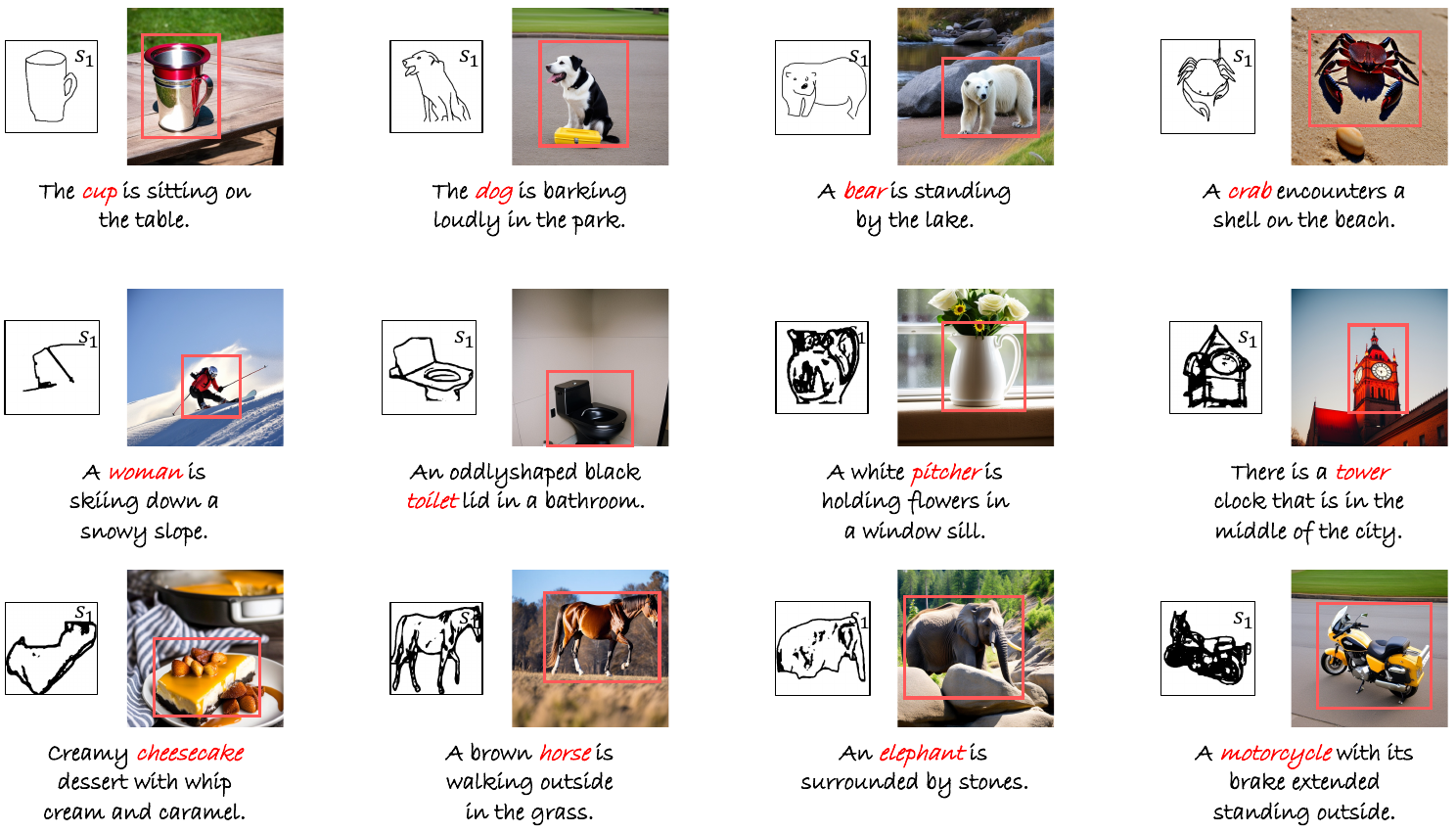}
    % \vspace{-0.3cm}
    \caption{The generation results of VersaGen when the user’s input of a single visual subject and text prompt are presented.}
    % \caption{VersaGen accommodates a variety of drawing inputs, enabling users to incorporate diverse forms of drawings such as single object drawing(first row), background drawing (last row), and more in conjunction with prompt.}
    \label{fig:add1}
\end{figure*}

\begin{figure*}[!h]
    \centering
    \includegraphics[width=0.98\textwidth]{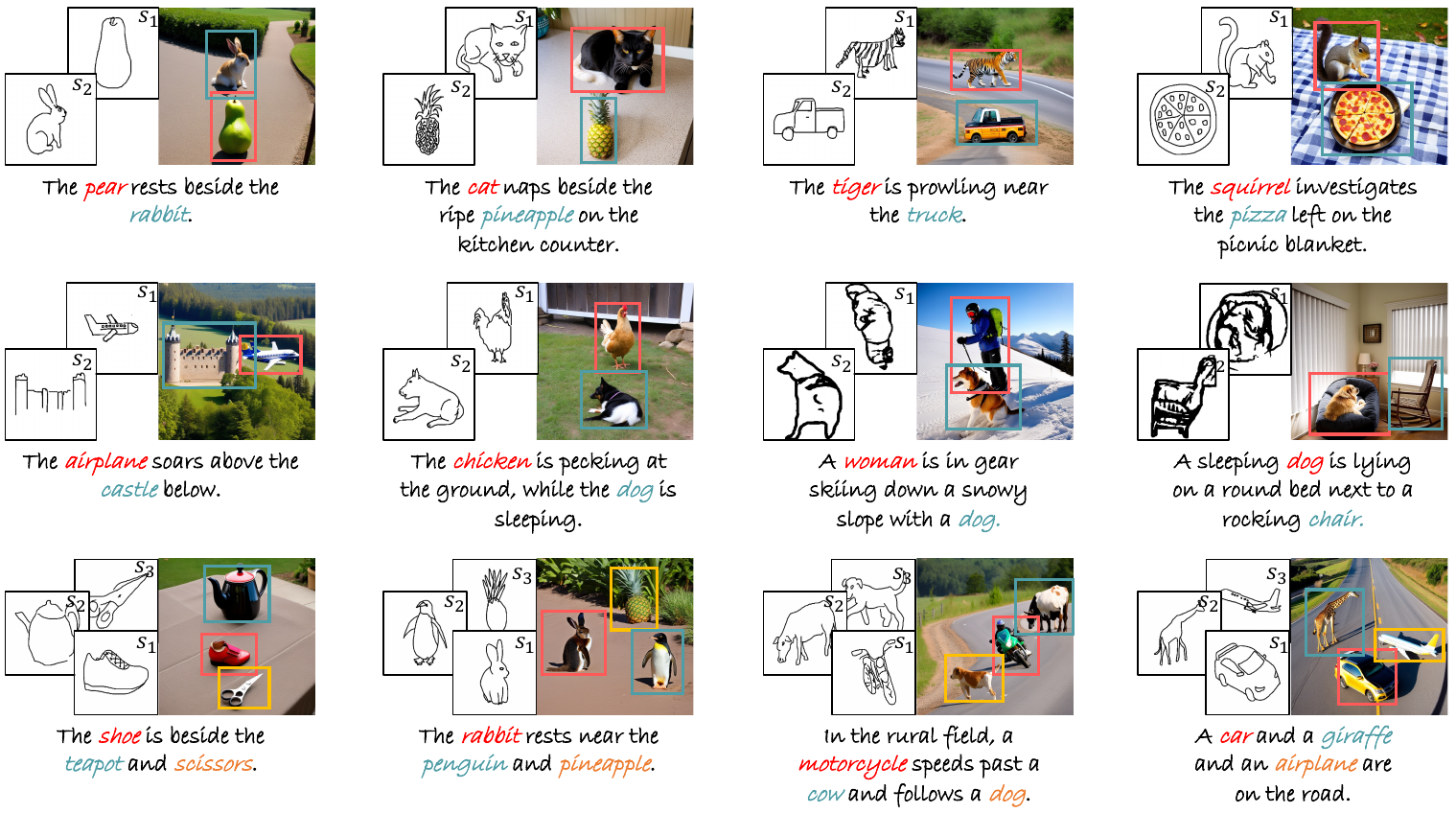}
    % \vspace{-0.3cm}
    \caption{The generation results of VersaGen when the user’s input of multiple visual subjects and text prompt are presented.}
    % \caption{VersaGen accommodates a variety of drawing inputs, enabling users to incorporate diverse forms of drawings such as single object drawing(first row), background drawing (last row), and more in conjunction with prompt.}
    \label{fig:add2}
\end{figure*}

\begin{figure*}[!h]
    \centering
    \includegraphics[width=0.98\textwidth]{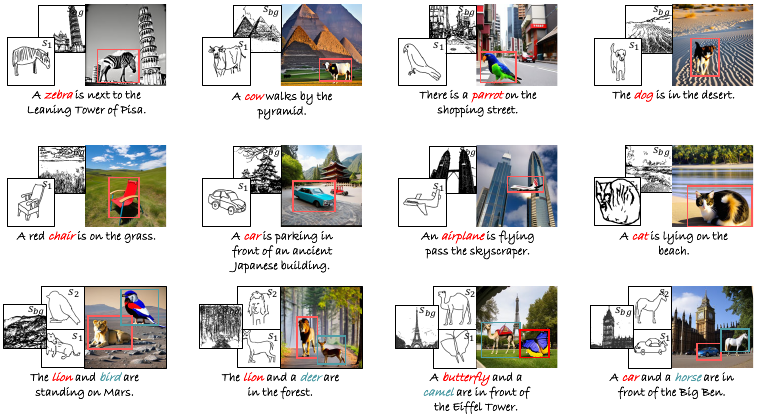}
    % \vspace{-0.3cm}
    \caption{The generation results of VersaGen when the user’s input of object-level drawings, a scene-level drawing and text prompt are presented.}
    % \caption{VersaGen accommodates a variety of drawing inputs, enabling users to incorporate diverse forms of drawings such as single object drawing(first row), background drawing (last row), and more in conjunction with prompt.}
    \label{fig:add3}
\end{figure*}

VersaGen is designed to handle a wide array of user inputs, including single visual subject (illustrated in Fig.~\ref{fig:add1}), multiple visual subjects (depicted in Fig.~\ref{fig:add2}), and combinations of both object-level and scene-level (presented in Fig.~\ref{fig:add3}).

\subsection{Discussion}
\subsubsection{Discussion on Scalability Challenges}
Due to the page limitation, we haven't discussed more about the scalability of VerseGen. Our approach, similar to ControlNet in design, is theoretically adaptable to various multimodal inputs (e.g., segmentation maps, OpenPose) and generative models (e.g., SDXL, SD3). It relies on the base generative model for subject localization, eliminating the need for external methods and ensuring scalability in terms of data and model diversity. Additionally, integrating user drawings as a control input strikes a balance between enhanced generation performance and maintaining user-friendly flexibility, supporting practical scalability across different applications.

\subsubsection{Discussion on Future Improvements}
Our method currently faces the following issues, which we plan to address in future improvements:

i) Support for Finer-Grained Level: We intend to expand our method to support the part of object inputs, aiming to achieve more precise control over image generation.

ii) Enhancing Reasoning Efficiency: We plan to integrate our method with other techniques to accelerate inference speeds, with the goal of further improving the efficiency of the generation process.

iii) Expanding Application Scope: We aim to apply our method to a broader range of modal conditions to validate its cross-modal applicability and versatility.

Through these improvements, we expect to further enhance the performance of our method and broaden its applications, thereby providing more possibilities for research and practice in the field of image generation.

\end{document}